%% file: main.tex
\documentclass[10pt,twocolumn,letterpaper]{article}

\usepackage[pagenumbers]{iccv} % To force page numbers, e.g. for an arXiv version

\input{preamble}

\definecolor{iccvblue}{rgb}{0.21,0.49,0.74}
\usepackage[pagebackref,breaklinks,colorlinks,allcolors=iccvblue]{hyperref}

\def\paperID{5775} % *** Enter the Paper ID here
\def\confName{ICCV}
\def\confYear{2025}

\title{\model: Learning Touch-Augmented Vision-Language-Action Models from Future Visual Supervision}

\author{
Ning Cheng\textsuperscript{\rm 1},
Jinan Xu\textsuperscript{\rm 1},
Wanlin Li\textsuperscript{\rm 2},
Yangzhi Chen\textsuperscript{\rm 1}, 
Jing Gao\textsuperscript{\rm 1}, 
Yiqun Wang\textsuperscript{\rm 1}, 
Kelan Peng\textsuperscript{\rm 1},
Wenjuan Han\textsuperscript{\rm 1}
\\
\small
 \textsuperscript{\rm 1}Beijing Jiaotong University~\textsuperscript{\rm 2}Beijing Institute for General Artificial Intelligence
}

\begin{document}

\maketitle

\input{sec/0_abstract}    
\input{sec/1_intro}
\input{sec/2_related_work}
\input{sec/3_method}
\input{sec/4_experiment}
\input{sec/5_conclusion}

{
    \small
    \bibliographystyle{ieeenat_fullname}
    \bibliography{main}
}

\end{document}

%% file: preamble.tex
\usepackage{placeins}

\usepackage{xspace}
\usepackage{pifont}
\usepackage{tcolorbox}
\usepackage{tablefootnote}
\usepackage{float}

\usepackage{multirow}
\usepackage{makecell}

\usepackage[linesnumbered,ruled,vlined]{algorithm2e}
\DontPrintSemicolon

\newcommand{\model}{$\tau$\xspace}
\newcommand{\dataset}{TacAura\xspace}

\definecolor{myblue}{RGB}{0,105,252}
\definecolor{myred}{RGB}{247,96,102}
\definecolor{mygray}{gray}{0.4}
\definecolor{red}{RGB}{255, 0, 0}
\definecolor{green}{RGB}{0, 100, 0}
\definecolor{gold}{RGB}{255, 125, 0}

\SetKwProg{For}{for}{:}{}

%% file: sec/0_abstract.tex
\begin{abstract}
Incorporating tactile sensing into Vision-Language-Action (VLA) models holds promise for contact-rich manipulation, where visual observations alone often fail to capture critical cues about physical interactions. However, learning informative tactile representation while effectively adapting it to pretrained VLA models remains challenging under limited task-specific data. Existing methods either focus on instantaneous contact states or model temporal interaction dynamics using 6D wrench sequences, leaving high-dimensional tactile signals underexplored.
To address these challenges, we present \model, a \textbf{t}ouch-\textbf{au}gmented VLA framework that learns an action-conditioned spatiotemporal tactile representation from future visual supervision inspired by the Joint-Embedding Predictive Architecture (JEPA), and fuses it with vision-language features for action generation. This supervision operates in latent space and is used only during training, adding no deployment overhead. We also introduce \dataset, a dataset of synchronized vision, proprioception, and vision-based tactile signals across four representative contact-rich manipulation tasks. Experiments show that \model outperforms existing models and generalizes to unseen objects and scenes, delivering improved manipulation performance and robustness. Project Page: \url{https://cocacola-lab.github.io/tau-Page/}.
\end{abstract}

%% file: sec/1_intro.tex
\section{Introduction}
\label{sec:intro}

%-------------------------------------------------------------------------
Recent advances \cite{zitkovich2023rt,kim2025openvla,black2024pi_0,black2025pi_} in Vision-Language-Action (VLA) models have established a new paradigm for general-purpose robot manipulation by leveraging large-scale multimodal pretraining across diverse tasks and embodiments. Nevertheless, their reliance on vision as the primary modality limits their capability in contact-rich manipulation, where successful execution depends on capturing physical interaction dynamics. Tactile sensing naturally provides this information, such as contact states, force distribution, and deformation patterns, which are difficult to infer from visual observations alone.

Despite the promise of tactile sensing, it remains nontrivial to learn tactile representations that capture interaction dynamics and effectively integrate them with the vision-language features of pretrained VLA models for action generation. Existing methods primarily incorporate touch through specialized fusion architectures \cite{yu2025forcevla,huang2025tactile,bi2026vla}, synchronous alignment with visual semantics \cite{cheng2026omnivtla} or force measurements \cite{huang2026tactile}, external tactile modules for action refinement \cite{zhang2026touchguide,liu2026taco}, or large-scale tactile training \cite{yuan2026ftp,niu2026t}. Despite these advances, these methods mainly focus on instantaneous contact states. Even when interaction dynamics are considered, it is typically modeled by 6D wrench sequences that capture temporal variations in structure-transmitted force–torque but lack dynamics-aware representation for high-dimensional vision-based tactile signals with fine-grained spatial contact patterns, whether by modeling their own temporal evolution or cross-modal predictive supervision \cite{zhang2026forceflow,niu2026t}. Therefore, how to model the temporal interaction dynamics for such high-dimensional signals under limited Vision-Tactile-Language-Action (VTLA) data remains an open problem.

To address the problem, we introduce \model, a \textbf{t}ouch-\textbf{au}gmented VLA framework that learns action-conditioned dynamics-aware tactile representations while preserving the capabilities of the pretrained VLA. Specifically, building upon the VLA backbone, \model introduces a tactile encoding and adaptation module that maps vision-based tactile signals into representations compatible with the semantic space of VLA. To encourage these representations to capture interaction dynamics, we further introduce an auxiliary predictive branch for cross-modal Self-Supervision Learning (SSL), inspired by Joint-Embedding Predictive Architecture (JEPA) \cite{lecun2022path,assran2023self}. Distinct from JEPA, which predicts target representations from contextual observations for general representation learning, our branch predicts future visual feature changes from the action-conditioned tactile representation to learn control-relevant interaction dynamics. Conditioned on the current tactile representation and the subsequent actions, a predictor forecasts the resulting change in future visual features. The target feature change is derived from the current and future observations encoded by the VLA vision encoder. A semantic similarity objective aligns the predicted feature change with this detached target, encouraging the tactile module to learn dynamics-aware representations that are predictive of future visual evolution under subsequent actions. Additionally, the predictive branch is used only during training, thereby improving representation quality without increasing inference complexity.

\begin{figure*}[t]
%\vskip -0.2in
\begin{center}
\centerline{\includegraphics[width=\linewidth]{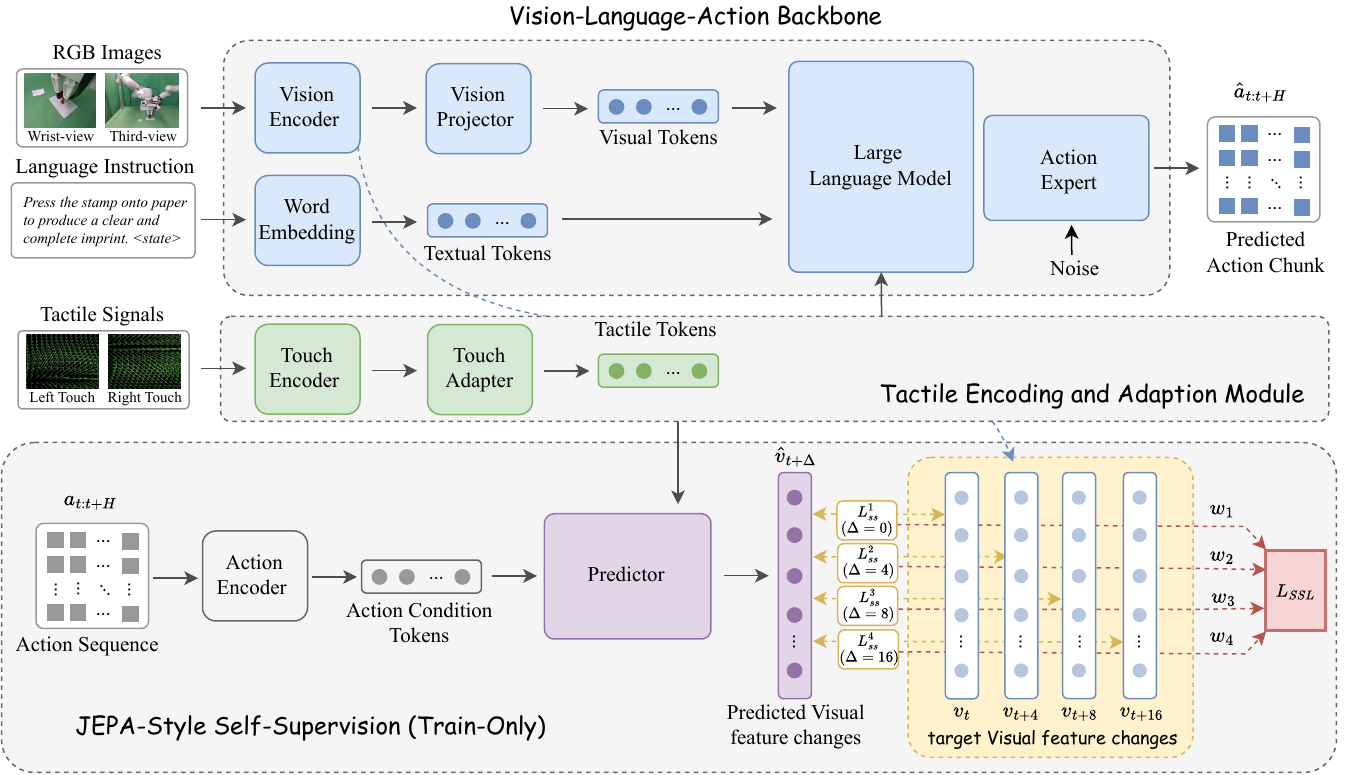}}
\caption{\textbf{\model Framework.} Multi-view visual observations, tactile signals, and language instruction are encoded into modality-specific tokens and fused by the large language model in the vision-language-action model for action chunk prediction. During training, an auxiliary JEPA-style self-supervised objective predicts future visual representations from action-conditioned latent features, enabling predictive multimodal representation learning without pixel-level reconstruction. Notably, the JEPA-style self-supervised learning is applied only during training and is removed during inference.}
\label{fig:tau_framework}
\end{center}
\vskip -0.2in
\end{figure*}

We evaluate \model with state-of-the-art visual-only VLA baselines and multimodal policies with tactile perception on multiple representative manipulation tasks requiring fine-grained physical interaction, including plug insertion, USB insertion, stamp press, and whiteboard erasing. The results show that \model consistently outperforms both vision-only VLA baselines and existing tactile-aware multimodal policies. Evaluations on unseen objects and scene configurations further demonstrate the model’s generalizability. In summary, our main contributions are as follows:
\begin{itemize}
    \item \textit{VTLA Framework.} We propose \model, a \textbf{t}ouch-\textbf{au}gmented VLA framework that learns dynamics-aware tactile representations from future visual supervision via action-conditioned prediction and integrates them into a pretrained VLA for action generation without additional inference overhead.
    \item \textit{Data infrastructure.} We establish a complete data infrastructure for contact-rich manipulation, including teleoperation tools, data conversion utilities, and a new dataset \dataset, enabling efficient collection and processing of VTLA robot data.
    \item \textit{Real-Robot Experiments.} Extensive real-robot experiments on four contact-rich manipulation tasks show that \model surpasses VLA and VTLA baselines and generalizes to unseen objects and scene variations.
\end{itemize}

%% file: sec/2_related_work.tex
\section{Related Work}
\label{sec:work}

\subsection{Tactile-Aware Robotic Manipulation}
Tactile-aware robotic manipulation aims to leverage tactile information for safe and reliable task execution. Some studies \cite{xue2025reactive,zhu2026touch,vitacmotor2026} build on lightweight policy models such as ACT \cite{zhao2023learning} and Diffusion Policy \cite{chi2023diffusion}, which are efficient to train and deploy but prone to overfitting task-specific datasets. Others \cite{zhang2026vtla} employ VLMs for tactile semantic understanding, cross-modal alignment, and robot action prediction. Still others \cite{ye2025learning,yuan2026vtam,lou2026dream} adopt world models to explicitly predict future tactile observations for learning interaction dynamics, but incur substantial computational overhead. Additionally, several approaches advance tactile-aware VLA modeling by either integrating tactile sensing into pretrained VLA backbones---including $\pi_0$ \cite{black2024pi_0}, $\pi_{0.5}$ \cite{black2025pi_}, and OpenVLA \cite{kim2025openvla}---with manipulation priors learned from large-scale robotic data \cite{yu2025forcevla,bi2026vla,huang2025tactile,cheng2026omnivtla,zhang2026touchguide,liu2026taco}, or developing VLA-style tactile foundation policies through large-scale multisensory pretraining \cite{liu2025mla,yuan2026ftp,niu2026t}. In this work, we investigate touch-augmented policy learning within the VLA paradigm under limited data and compute budgets by learning the tactile representations that capture instantaneous contact states and temporal interaction dynamics, and fusing it with visual-language features to better ground action generation.

\subsection{Cross-Modal Self-Supervision}
Cross-modal self-supervision leverages the natural correspondences among different sensory modalities to learn effective representations without manual annotations. Early work \cite{socher2013zero,owens2016ambient,lu2019vilbert} mainly focuses on bimodal or trimodal representation learning across vision, language, and audio. Building on these efforts, subsequent studies extend this approach to visual-tactile learning by leveraging the synchronization and temporal correlations between visual and tactile signals. Zambelli \etal. \cite{zambelli2021learning} explore several cross-modal self-supervised objectives for visual-tactile learning, showing that visual signals can provide effective supervision for tactile representations. Yang \etal. \cite{yang2022touch} learn tactile representations
for the Gelsight sensor with visuo-tactile contrastive multiview
coding \cite{tian2020contrastive}. Kerr \etal. \cite{kerr2022self}, Yang \etal.  \cite{yang2024binding}, Cheng \etal.  \cite{cheng2025touch100k} propose various contrastive pretraining methods. 
Cao \etal. \cite{cao2023learn} apply masked autoencoders to learn tactile representations directly from tactile inputs. Feng \etal. \cite{feng2025anytouch,feng2026anytouch} exploit masked autoencoders and contrastive learning. Unlike these methods, we introduce JEPA-style self-supervision into touch-augmented VLA policy learning, leveraging correspondences between conditioned actions and future visual feature changes to learn dynamics-aware tactile representations, thereby enabling the policy to more effectively integrate interaction cues with vision-language features and generate contact-aware actions.

%% file: sec/3_method.tex
\section{The \model Framework}
In this section, we detail the \model's
model architecture, training strategy, and training dataset.

\subsection{Model Architecture}
As illustrated in Figure \ref{fig:tau_framework}, \model consists of three key components: (1) a pretrained $\pi_{0.5}$ Vision-Language-Action (VLA) backbone for multimodal perception and action generation, (2) a tactile encoding and adaptation module that bridges tactile signals with the latent space of the pretrained VLA model, and (3) an auxiliary JEPA-style predictive self-supervised branch that facilitates tactile representation learning through future visual latent prediction.

\paragraph{Pretrained $\pi_0.5$ VLA Backbone.}
\model is built upon the pretrained VLA model $\pi_{0.5}$, which serves as the foundation for multimodal policy learning. At each time step $t$, the VLA model receives an observation $o_t=\{\mathcal{I}_t,\tilde{\ell}_t\}$,
where $\mathcal{I}_t=\{\mathcal{I}_t^1,\ldots,\mathcal{I}_t^N\}$ denotes the set of $N$ RGB images, and
$\tilde{\ell}_t=[\ell_t;q_t]$
represents the language instruction formed by concatenating the task description $\ell_t$ with the robot's proprioceptive state $q_t$. After encoding and fusing these multimodal inputs, the action expert predicts a future action chunk $\hat{a}_{t:t+H}$ from noise via conditional flow matching.

\paragraph{Tactile Encoding and Adaptation Module.}
To endow \model with tactile perception capability, we introduce a tactile encoding and adaptation module, which transforms raw tactile signals into latent representations compatible with the multimodal embedding space of the VLA model. Specifically, the observation at time step $t$ is extended from $o_t=\{\mathcal{I}_t,\tilde{\ell}_t\}$ to $\tilde{o}_t=\{\mathcal{I}_t,\tilde{\ell}_t,\mathcal{T}_t\}$, where $\mathcal{T}_t=\{\mathcal{T}_t^{L},\mathcal{T}_t^{R}\}$ denotes the tactile signals acquired from the left and right tactile sensors mounted on the robot gripper, each formed by concatenating separately normalized single-channel normal and two-channel shear maps. Similar to the vision encoder that transforms RGB images into visual tokens, the tactile signals are first processed by a touch encoder $E_{\text{tou}}$ to extract tactile features:
\begin{equation}  
z_t^{\mathrm{touch}}
=
\left[E_{\mathrm{tou}}(\mathcal{T}_t^L);E_{\mathrm{tou}}(\mathcal{T}_t^R)\right],
\end{equation}
where \([\,;\,]\) denotes token-wise concatenation, $E_{\text{tou}}$ is shared by the left and right tactile signals and initialized from the $\pi_{0.5}$ vision encoder, and $z_t^{\text{touch}}$ encodes bilateral normal and shear deformation features.

Since the pretrained VLA backbone operates in the multimodal embedding space learned from visual and textual data, the extracted tactile features cannot be directly integrated with the existing token representations. Therefore, we introduce a learnable linear adapter $A_{\text{tou}}$ to project the tactile representations into the latent space of the pretrained VLA model:
\begin{equation}  
\mathcal{Z}_t^{\text{touch}}=A_{\text{tou}}(z_t^{\text{touch}}),
\end{equation}
where $\mathcal{Z}_t^{\text{touch}}$ is the LLM-aligned tactile token sequence. 

Subsequently, the adapted tactile tokens are concatenated with the visual tokens and textual tokens form a unified token-level semantic representation $\mathcal{Z}_t=[\mathcal{Z}_t^{\text{vision}};\mathcal{Z}_t^{\text{language}};\mathcal{Z}_t^{\text{touch}}]$, which is processed by the large language model and then is seen by the action expert to help predict the future action chunk $\hat{a}_{t:t+H}$. The tactile tokens are also fed into the auxiliary predictive branch during training.

\paragraph{JEPA-style Predictive Self-Supervised Branch.} 
Besides policy learning, we present an auxiliary JEPA-style predictive self-supervised branch to encourage tactile representations to encode how the visually observed interaction state evolves under subsequent actions. Specifically, the adapted tactile tokens $\mathcal{Z}_t^{\text{touch}}$ from the tactile encoding and adaptation module, together with the corresponding action tokens $\mathcal{Z}_{t:t+H}^{\text{action}}$ are concatenated and fed into a predictor network $P(\cdot)$ to predict the future visual latent representations:
\begin{equation} 
\{\hat{z}_{t+\Delta_{k}}^{\text{vision}}\}_{k=1}^K =
P(\mathcal{Z}_t^{\text{touch}}, \mathcal{Z}_{t:t+H}^{\text{action}}),
\end{equation}
where $\Delta_k$ denotes a predefined temporal offset relative to the current time step $t$. The predictor $P(\cdot)$ adopts a lightweight multilayer perceptron (MLP) architecture with three fully connected layers and GELU activations. The action tokens $\mathcal{Z}_{t:t+H}^{\text{action}}$ are obtained by first vectorizing the action sequence $a_{t:t+H}$ and encoding it with an action encoder $E_{\text{act}}$, implemented as a MLP with two linear layers:
\begin{equation} 
\mathcal{Z}_{t:t+H}^{\text{action}} = E_{\text{act}}(vec(a_{t:t+H})),
\end{equation}
The prediction target is the latent representation of the future visual observation, which is extracted by the vision encoder $E_{\text{vis}}$ of the pretrained VLA backbone:
\begin{equation} 
\{{z}_{t+\Delta_{k}}^{\text{vision}}\}_{k=1}^K 
=
E_{\text{vis}}
(\{\mathcal{I}_{t+\Delta_k}\}_{k=1}^K),
\end{equation}
where $\mathcal{I}_{t+\Delta_k}$ denotes a set of the future RGB images at the temporal offset $\Delta_k$.

\subsection{Training Strategy}
The proposed \model is trained end-to-end by jointly optimizing a supervised imitation learning objective and a predictive self-supervised objective. The former enables the pretrained VLA policy to incorporate tactile information for action generation, while the latter provides complementary supervision for learning more informative physical interaction dynamics.

\paragraph{Action Policy Adaptation for VTLA.} 
To enable the pretrained VLA policy to leverage tactile information for action generation, we adapt it through supervised imitation learning on expert demonstrations. From each demonstration trajectory consisting of vision-language-touch observations and expert actions $\{(\tilde{o}_t,a_t)\}_{t=1}^{T}$, we construct each training sample by pairing the observation at time step $t$ with the subsequent expert action sequences $a_{t:t+H}$ as the supervision target. Following the conditional flow-matching for action modeling \cite{black2024pi_0, black2025pi_}, a random Gaussian noise $\epsilon\sim\mathcal{N}(0,I)$ is sampled, and the noisy actions are constructed as $a_{t:t+H}^s=s\cdot a_{t:t+H}+(1-s)\cdot\epsilon$, where $s\in[0,1]$ represents the continuous timestep along the flow trajectory. The denoising directions along the flow trajectory are obtained as the derivative of the noisy actions $a_{t:t+H}^s$ with respect to the flow timestep $s$:
\begin{equation} 
u(a_{t:t+H}^s | a_{t:t+H}) = \frac{\partial a^{s}_{t:t+H}}{\partial s} =
a_{t:t+H}-\epsilon.
\end{equation} 
The action expert learns a denoising vector field $v_\theta(a_{t:t+H}^s,\tilde{o}_t)$ to match this target direction $u(a_{t:t+H}^s | a_{t:t+H})$, resulting in the following policy objective:
\begin{equation} 
\mathcal{L}_{\mathrm{IL}}
= 
\mathbb{E}
\left[
\left\|
v_\theta(a_{t:t+H}^s,\tilde{o}_t)
-
u(a_{t:t+H}^s | a_{t:t+H})
\right\|_2^2
\right].
\end{equation} 
where the expectation is taken over the expert action distribution \(p(a_{t:t+H}|\tilde{o}_t)\) and the conditional flow path \(q(a_{t:t+H}^{s}|a_{t:t+H})\).

\paragraph{Joint Fine-tuning with Predictive Self-Supervision.}
To encourage tactile representations to capture interaction dynamics rather than static information, instead of directly supervising absolute visual features, the auxiliary objective supervises future visual feature changes relative to the current observation. The predicted feature variations and their detached target are defined as:
$
\Delta \hat{z}^{\mathrm{vision}}_{t+\Delta_k}
=
\hat{z}^{\mathrm{vision}}_{t+\Delta_k}
-
z^{\mathrm{vision}}_t$, $
\Delta z^{\mathrm{vision}}_{t+\Delta_k}
=
\operatorname{sg}(z^{\mathrm{vision}}_{t+\Delta_k}
-
z^{\mathrm{vision}}_t)$, where $\operatorname{sg}$ denotes stop-gradient.
To emphasize future task processes with significant contact transitions, each future prediction is weighted according to the tactile variation magnitude $w_{t+\Delta_k}$:
\begin{equation}
w_{t+\Delta_k}
=
\underset{i\in\{L,R\}}{\operatorname{mean}}
\operatorname{clip}\!\left(
\frac{
\left\lVert
\mathcal{T}_{t+\Delta_k}^{(i)}
-
\mathcal{T}_{t}^{(i)}
\right\rVert_2
}{
\bar{c}+\delta
},
w_{\min},
w_{\max}
\right).
\end{equation}
where \(i\in\{L,R\}\) indexes the two tactile sensors, \(\bar c\) is the normalization scale from the training set, and $\delta=1e-6$ ensures numerical stability. We set \(w_{\min}=0.2\) and \(w_{\max}=5.0\) to limit the effects of negligible and abnormal tactile variations.
Based on the tactile-derived weights $\{w_{t+\Delta_{k}}\}_{k=1}^K$, the predictive self-supervised objective is formulated as a weighted cosine alignment loss:

\begin{equation}
\mathcal{L}_{\mathrm{SSL}}
=
\frac{
\sum_{k=1}^{K}
w_{t+\Delta_k}
\left(
1-
\cos
(
\Delta\hat z^{\mathrm{vision}}_{t+\Delta_k},
\Delta z^{\mathrm{vision}}_{t+\Delta_k}
)
\right)
}
{
\sum_{k=1}^{K}w_{t+\Delta_k}+\delta
}.
\end{equation}
The final training objective jointly optimizes the imitation learning objective and the predictive self-supervised objective:
\begin{equation}
\mathcal{L}_{\mathrm{total}}
=
\mathcal{L}_{\mathrm{IL}}
+
\lambda\mathcal{L}_{\mathrm{SSL}} .
\end{equation}
where $\lambda$ balances the contribution and is set to 0.3.
% of tactile representation learning.

\begin{figure*}[htp]
\vskip -0.2in
\begin{center}
\centerline{\includegraphics[width=\linewidth]{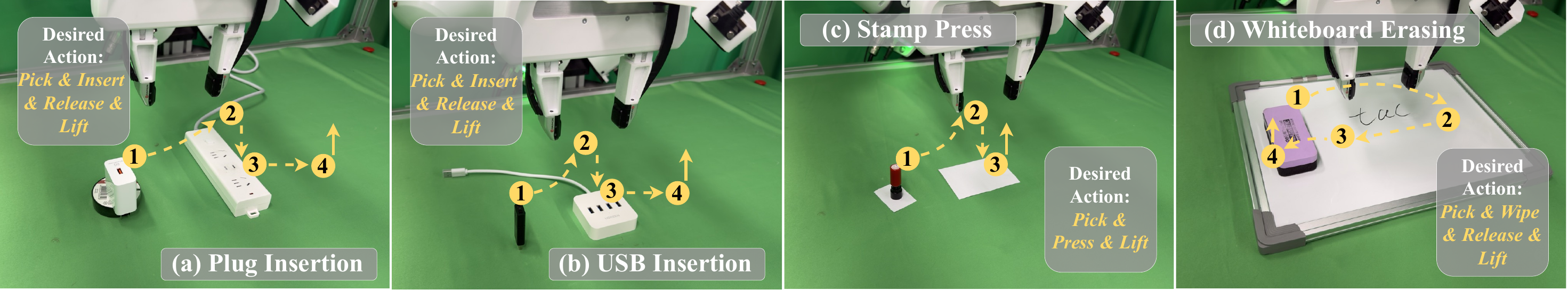}}
\caption{Contact-rich task suite of \dataset. 
}
\label{fig:task_suite}
\end{center}
\vskip -0.2in
\end{figure*}

\subsection{\dataset Dataset}
\label{sec:dataset}

A central obstacle to VTLA learning lies not only in the model but also in the data infrastructure. We thus develop a lightweight infrastructure comprising teleoperation tools, multi-stream synchronization and conversion utilities, and the \dataset dataset, all of which will be open-sourced. In this subsection, we focus on the \dataset dataset, while details of the remaining data infrastructure components are provided in the appendix.

\dataset is a tactile-aware manipulation suite comprising four contact-rich tasks with complementary interaction characteristics, as illustrated in Figure \ref{fig:task_suite}. Plug insertion and USB insertion require precise geometric alignment under tight tolerances, whereas stamp pressing and whiteboard erasing emphasize contact-force sensing and modulation under less restrictive geometric constraints.

The data collection is performed using a Franka Research 3 arm equipped with a Franka Hand end-effector, where the gripper fingers are replaced by DM-Tac WS, a vision-based tactile sensing system that provides high-resolution tactile feedback (320x240, $\sim$ 40 FPS). The sensors are mounted onto the hand using a custom 3D-printed adapter. Visual observations are captured from three RGB-D cameras: two static third-person views (RealSense D435i at 640x480, 15 FPS) and one wrist-mounted camera (RealSense D405 at 640x480, 15 FPS) providing egocentric perspectives. Expert demonstrations are collected via human teleoperation with visualized tactile feedback using an exoskeleton arm kinematically isomorphic to the Franka Research 3 arm, mapping to robot joint position control. 

All device streams are recorded synchronously during data collection. Since different devices may have different startup times, we further align all streams based on their timestamps, downsample the data to 10 Hz, and annotate each trajectory with the corresponding task description, as detailed in the appendix. As a result, this process yields 100 high-quality human demonstrations with varied object poses for each task.

%% file: sec/4_experiment.tex
%%%%%%%%%%%%%%%%%%%%%
%%%%%% Main Results
\begin{table*}[!t]
%\vskip -0.2in
\centering
\resizebox{\textwidth}{!}{
\begin{small}
\begin{tabular}{lcccccc}
\toprule
\multirow{2}{*}{\textbf{Model / Task}} & \textbf{Plug Insertion} & \textbf{USB Insertion} & 
\textbf{Stamp Press} &
\textbf{Whiteboard Erasing} & \multirow{2}{*}{\textbf{Avg.}} \\
\cmidrule(lr){2-2} \cmidrule(lr){3-3} \cmidrule(lr){4-4} \cmidrule(lr){5-5} 
& Grasp / Align / Insertion & Grasp / Align / Insertion & Pick / Contact / Stamp & Pick / Contact / Wipe \\ 
\midrule
$\pi_0$  \cite{black2024pi_0} & 100 / 25 / \hphantom{0}0 & 100 / 25 / 15  & 100 / \hphantom{0}65 / 30 & 100 / 100 / 35 & 20.00 \\ 
$\pi_{0.5}$ \cite{black2025pi_} & 100 / 50 / 20 & 100 / 35 /  20  &  100 / \hphantom{0}70 / 35  & 100 / 100 / 40 & 28.75 \\
 ForceVLA$^\dagger$  \cite{yu2025forcevla} & 100 / 85 / \hphantom{0}0 & 100 / 45 / \hphantom{0}5  & 100 / 100 / 70 & 100 / 100 / 45 & 30.00 \\ 
 ForceFlow$^\dagger$ \cite{zhang2026forceflow} & \hphantom{0}90 / 40 / \hphantom{0}0 & \hphantom{0}85 / 30 / \hphantom{0}0  & \hphantom{0}90 / \hphantom{0}60 / 45 & \hphantom{0}95 / \hphantom{0}95 / 50 & 23.75 \\
 T-Rex$^\dagger$  \cite{niu2026t} & 100 / 30 / 30 & \hphantom{0}95 / 30 / 30 & 100 / \hphantom{0}70 / 35 & 100 / 100 / 30 & 31.25 \\ 
 \midrule
 $\tau\text{-Wrist}^{Sup.}$ & 100 / 75 / \underline{60} & 100 / 50 / \underline{40}  &  100 / 100 / \textbf{90}   & 100 / 100 / \textbf{95} & \textbf{71.25}\\
$\tau\text{-Front}^{Sup.}$ & 100 / 80 / \textbf{65} & 100 / 70 / \textbf{50}  &  100 / 100 / \underline{75}   & 100 / 100 / \underline{85} & \underline{68.75} \\
$\tau\text{-DualView}^{Sup.}$ & 100 / 70 / 55 & 100 / 50 / 35  &  100 / 100 / 65   & 100 / 100 / 75 & 57.50\\
\bottomrule
\end{tabular}
\end{small}
}
\caption{\textbf{Success rates (\%) of different models on four contact-rich tasks.} The best results are in \textbf{bold}, and the suboptimal ones are \underline{underlined}. \textbf{Avg.} denotes the average of the full success rates across all four tasks. $^\dagger$ marks models adapted for our tactile sensing setup while preserving the original framework. $\tau\text{-Wrist}^{Sup.}$, $\tau\text{-Front}^{Sup.}$, and $\tau\text{-DualView}^{Sup.}$ are three variants of the \model framework that use wrist-view, front-view, and dual-view representations, respectively, as supervisory signals for SSL.} 
\label{tab:main-res}
%\vskip -0.2in
\end{table*}
%%%%%% Main Results
%%%%%%%%%%%%%%%%%%%%%

\begin{figure*}[t]
%\vskip -0.2in
\begin{center}
\centerline{\includegraphics[width=\linewidth]{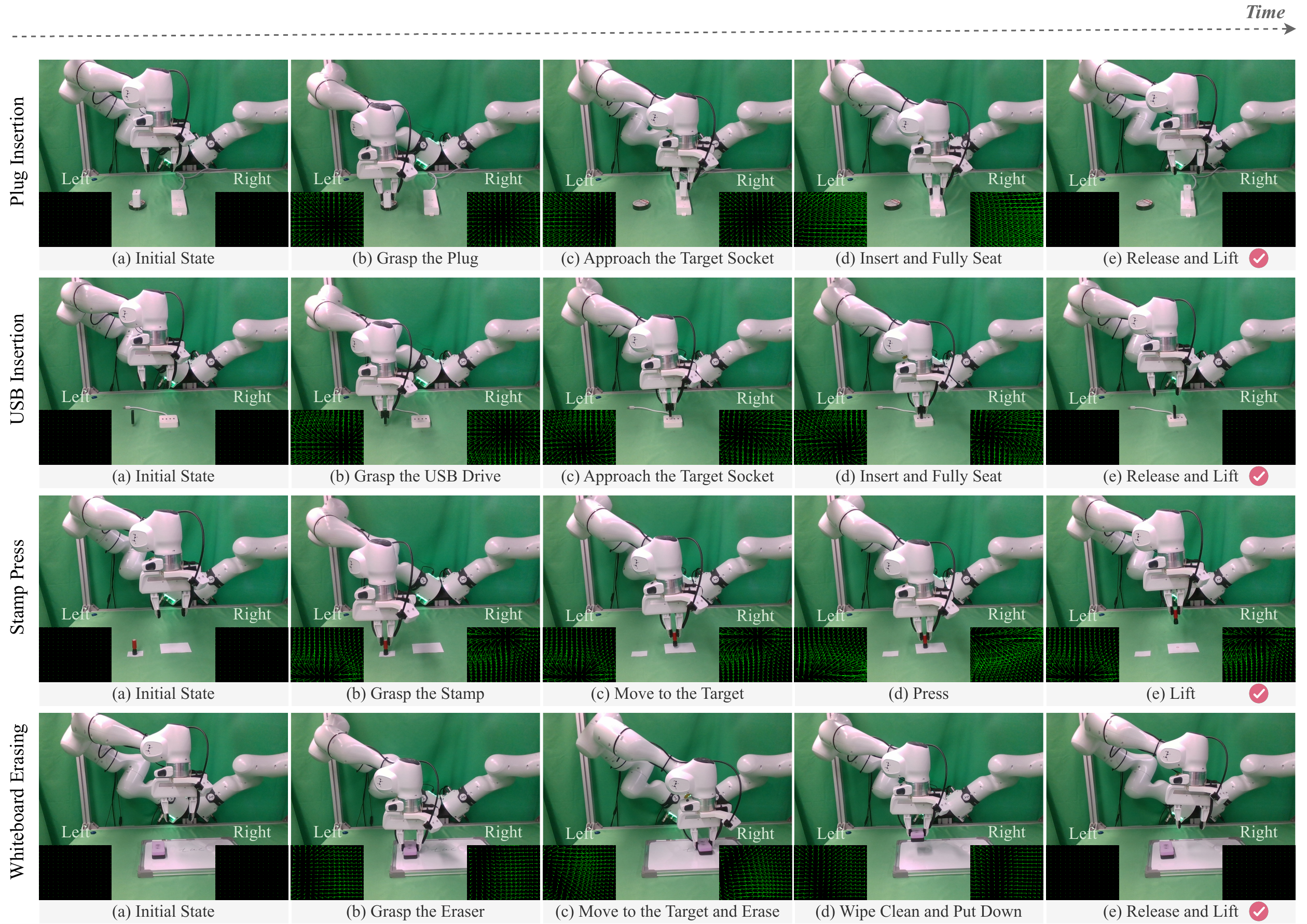}}
\caption{Qualitative results of policy execution.}
\label{fig:qualitative_res}
\end{center}
\vskip -0.2in
\end{figure*}

\begin{figure}[t]
\vskip -0.2in
\begin{center}
\includegraphics[width=\linewidth]{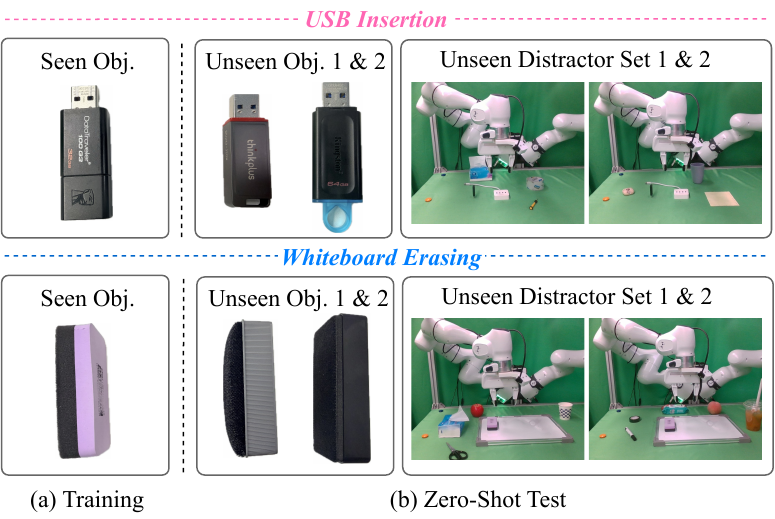}
\caption{Training and zero-shot testing settings for USB insertion and whiteboard erasing tasks.
%The models are trained on seen objects and tested on unseen objects and distractor sets to assess their generalization to novel objects and complex scenes.
}
\label{fig:general_show}
\end{center}
%\vskip -0.2in
\end{figure}

\begin{figure}[t]
\vskip -0.2in
\begin{center}
\includegraphics[width=\linewidth]{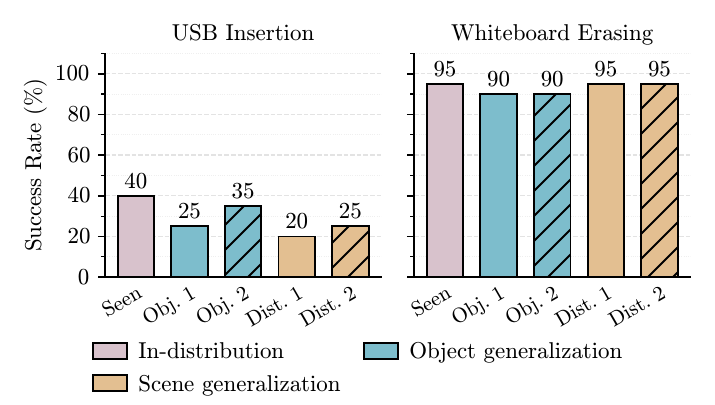}
\caption{Success rates for USB insertion and whiteboard erasing under in-distribution, object-generalization, and scene-generalization settings.}
\label{fig:generalization_res}
\end{center}
\vskip -0.2in
\end{figure}

\section{Experiments}
We train \model on \dataset dataset, and evaluate it in the real world across four tasks from TacAura: plug insertion, USB insertion, stamp press, and whiteboard erasing.

\subsection{Experimental Setup} 
 We train \model with 100 expert demonstrations per task for 30,000 steps, 
 %using a cosine learning-rate schedule with 10,000 warmup steps and a peak learning rate of $5e-5$,
and compare it against recent representative open-source approaches, including the general-purpose VLA models $\pi_0$ \cite{black2024pi_0} and $\pi_{0.5}$ \cite{black2025pi_}, as well as methods designed for contact-rich manipulation: ForceVLA$^\dagger$ \cite{yu2025forcevla}, ForceFlow$^\dagger$ \cite{zhang2026forceflow}, and T-Rex$^\dagger$ \cite{niu2026t}. Each model is evaluated over 20 trials under moderate randomization. The evaluation metric is the success rate at each task stage. For plug insertion and USB insertion, each task is divided into three stages: grasping (Grasp), aligning with the port and initiating insertion (Align), and fully inserting the object (Insertion). For stamp press and whiteboard erasing, each task is also divided into three stages: picking up the object (Pick), establishing effective contact (Contact), and producing a full stamp mark (Stamp)/completely erasing the target area (Wipe). Further experimental setup details are provided in the appendix.

\subsection{Main Results}
Table \ref{tab:main-res} compares the different models across the four tasks, and Figure \ref{fig:qualitative_res} visualizes the qualitative results of \model's policy execution. Among the baselines, T-Rex$^\dagger$ achieves the highest average full-task success rate of 31.25\%.
%, followed by $\pi_{0.5}$ (28.75\%), ForceFlow$^\dagger$ (23.75\%), and $\pi_0$ (20.00\%). 
In comparison, all three variants of the $\tau$ framework achieve substantially higher average success rates. Specifically, $\tau\text{-Wrist}^{Sup.}$ obtains the best overall performance with an average success rate of 71.25\%, outperforming the strongest baseline by 40 percentage points. $\tau\text{-Front}^{Sup.}$ achieves a comparable average of 68.75\%, whereas $\tau\text{-DualView}^{Sup.}$ reaches 57.50\%. These results demonstrate the considerable performance advantage of the \model framework.

The stage-wise examination in Table \ref{tab:main-res} shows that the primary limitation of the baseline models lies in final task completion rather than initial object acquisition or contact establishment. Most methods achieve high grasping, picking, or contact success rates, but the baselines often exhibit substantial degradation at the final execution stage. For example, ForceVLA$^\dagger$ achieves an 85\% alignment success rate on Plug Insertion but fails to complete any insertion, while ForceFlow$^\dagger$ reaches a 95\% contact success rate on Whiteboard Erasing but completes the wiping operation in only 50\% of the trials. In contrast, the $\tau$ variants retain substantially higher success rates from the intermediate stage to final completion. The best $\tau$ variant achieves full-task success rates of 65\%, 50\%, 90\%, and 95\% on plug insertion, USB insertion, stamp press, and whiteboard erasing, respectively. These results exceed the strongest baseline on the corresponding tasks by 35, 20, 20, and 45 percentage points, indicating that the proposed framework is particularly effective in improving fine-grained contact reasoning and execution.

The three \model variants reveal a task-dependent effect of the supervisory view. $\tau\text{-Front}^{Sup.}$ performs best on plug and USB insertion tasks, achieving 65\% and 50\% full-task success, respectively, suggesting that front-view features may provide more useful spatial cues for target-hole localization. By contrast, $\tau\text{-Wrist}^{Sup.}$ performs best on stamp press (90\%) and whiteboard erasing (95\%), where wrist-view observations are less occluded than in the other tasks and thus provide clearer local cues. Although $\tau\text{-DualView}^{Sup.}$ outperforms the baselines on average, it underperforms the best single-view variant and even falls slightly below ForceVLA$^\dagger$ on stamp press. This may be because learning complementary supervisory signals from two views is challenging with limited data and a simple fusion strategy, motivating a more effective view-fusion mechanism.

%%%%%%%%%%%%%%%%%%%%%
%%%%%% ab study
\begin{table*}[h!t]
\vskip -0.2in
\centering
\resizebox{\textwidth}{!}{
\begin{small}
\begin{tabular}{lcccccc}
\toprule
\multirow{2}{*}{\textbf{Model / Task}} & \textbf{Plug Insertion} & \textbf{USB Insertion} & 
\textbf{Stamp Press} &
\textbf{Whiteboard Erasing} & \multirow{2}{*}{\textbf{Avg.}} \\
\cmidrule(lr){2-2} \cmidrule(lr){3-3} \cmidrule(lr){4-4} \cmidrule(lr){5-5} 
& Grasp / Align / Insertion & Grasp / Align / Insertion & Pick / Contact / Press & Pick / Contact / Wipe \\ 
\midrule
 \model (Ours)   & 100 / 75 / \textbf{60} & 100 / 50 / \textbf{40} & 100 / 100 / \textbf{90} & 100 / 100 / \textbf{95} & \textbf{71.25} \\ 
 \quad w/o Action Seq. & 100 / 75 / 55 & 100 / 45 / 30  &  100 / 100 / 75   & 100 / 100 / 75 & 58.75 \\
 \quad w/o SSL & 100 / 75 / 50 & 100 / 50 / 25  &  100 / 100 / 70   & 100 / 100 / 60 & 51.25 \\
 \quad w/o Tactile Module & 100 / 50 / 20 & 100 / 35 / 20  & 100 / \hphantom{0}70 / 35 & 100 / 100 / 40 & 28.75 \\ 
 
\bottomrule
\end{tabular}
\end{small}
}
\caption{Ablation study following an incremental removal protocol on all four tasks. \textit{w/o Tactile Module} corresponds to $\pi_{0.5}$.} 
\label{tab:ab-res}
%\vskip -0.2in
\end{table*}
%%%%%% ab study
%%%%%%%%%%%%%%%%%%%%%

\subsection{Generalization Analysis}

To evaluate the zero-shot generalization capability of \model, we consider USB insertion and whiteboard erasing, training the models and evaluating them on two unseen objects and two unseen distractor sets for each task, as shown in Figure \ref{fig:general_show}. The unseen objects differ from the training objects in visual appearance and surface texture, while the distractor sets introduce additional objects and visual clutter into the workspace. These settings assess object-level and scene-level generalization, respectively.

\paragraph{Object Generalization.} As shown in Figure \ref{fig:generalization_res}, the success rate of USB insertion decreases from 40\% on the seen object to 25\% and 35\% on the two unseen USB drives, yielding an average object-generalization success rate of 30\%. The different performance across the two unseen instances suggests that USB insertion remains sensitive to variations in object geometry and contact configuration. Nevertheless, the model retains meaningful zero-shot insertion capability. Whiteboard Erasing exhibits substantially stronger object generalization: the success rate decreases only slightly from 95\% on the seen eraser to 90\% on both unseen erasers. This limited performance gap indicates that the learned representation transfers effectively across erasers with different appearances and textures.

\paragraph{Scene Generalization.} The results reveal a clear task-dependent difference in scene generalization. For USB Insertion, the success rate decreases from 40\% in the in-distribution setting to 20\% and 25\% under the two unseen distractor sets, respectively. This corresponds to an average success rate of 22.5\%, representing a 17.5-percentage-point drop from the in-distribution result. The degradation is also larger than that observed under unseen-object variations, indicating that visual clutter poses a substantial challenge to target localization and precise alignment in USB insertion. In contrast, whiteboard erasing maintains a 95\% success rate under both distractor configurations, showing no degradation relative to the in-distribution setting.

\subsection{Ablation Study}
To investigate the impact of key components in \(\tau\), we ablate the action sequence conditioning, predictive self-supervised learning, and tactile encoding and adaptation module on the best-performing $\tau\text{-Wrist}^{Sup.}$, with similar trends observed for the other variants. Results are shown in Table \ref{tab:ab-res}.

\paragraph{Impact of Action Sequence Conditioning.} Removing action sequence conditioning reduces the average full-task success rate from 71.25\% to 58.75\%, with drops of 5, 10, 15, and 20 percentage points on plug Insertion, USB insertion, stamp press, and whiteboard erasing tasks, respectively. Initial grasping, picking, and contact stages remain largely unchanged, indicating that action sequence conditioning primarily contributes to final execution by capturing temporal dependencies among consecutive actions. Its larger impact on stamp press and whiteboard erasing further highlights that coherent action context is particularly important for tasks that rely more heavily on precise force control.

\paragraph{Impact of Predictive Self-Supervised Learning.} Removing predictive self-supervised learning decreases the average full-task success rate from 71.25\% to 51.25\%, resulting in a 20-point drop. The full-task success rates decrease by 10, 15, 20, and 35 percentage points on plug insertion, USB insertion, stamp press, and whiteboard erasing, respectively, while the intermediate alignment and contact success rates remain unchanged. This phenomenon suggests that the predictive objective helps the model capture the temporal evolution for tactile representations during sustained contact. These results demonstrate that predictive self-supervised learning improves the quality of the learned representations and enables more reliable contact-aware execution.

\paragraph{Impact of Tactile Encoding and Adaptation Module.} Removing the tactile encoding and adaptation module leads to the largest performance degradation, reducing the average full-task success rate from 71.25\% to 28.75\%, a decrease of 42.50 percentage points. The full-task success rates on the four tasks drop by 40, 20, 55, and 55 percentage points, respectively. Unlike the other two ablations, removing this module also substantially reduces intermediate-stage performance, including alignment on the insertion tasks and contact establishment on stamp press. Meanwhile, grasping and picking success rates remain at 100\%, indicating that coarse object interaction can still be achieved without touch, whereas precise contact reasoning and execution cannot. 

%% file: sec/5_conclusion.tex
\section{Conclusion}
In this paper, we presented \model, a touch-augmented vision-language-action framework that extends pretrained VLA models with tactile perception for contact-rich robotic manipulation. By introducing a tactile encoding and adaptation module, \model effectively aligns tactile representations with the latent embedding space of the pretrained VLA backbone, enabling data-efficient multimodal fusion for action generation. 

Furthermore, we proposed an auxiliary JEPA-style predictive self-supervised branch that learns temporally consistent multimodal representations through latent future prediction. The predictive objective provides auxiliary supervision during training and is removed during inference, resulting in improved representation quality without additional deployment cost. Extensive experiments demonstrate that the proposed framework improves manipulation performance and robustness across representative contact-rich tasks, while exhibiting generalization to unseen objects and scenes. 

Although \model achieves strong performance, computational resource constraints prevent comparisons with tactile-aware world models. Moreover, its cross-task, cross-embodiment, and cross-sensor transferability remains unexplored and will be investigated in future work.